\documentclass[10pt,twocolumn,letterpaper]{article}

\usepackage{iccv}
\usepackage{times}
\usepackage{epsfig}
\usepackage{graphicx}
\usepackage{amsmath}
\usepackage{amssymb}
\usepackage{subcaption} 

\usepackage[pagebackref=true,breaklinks=true,letterpaper=true,colorlinks,bookmarks=false]{hyperref}

\iccvfinalcopy 

\def\iccvPaperID{44} 

\ificcvfinal\fi

\begin{document}

\title{A Possible Reason for why Data-Driven Beats Theory-Driven Computer Vision}

\author{John K. Tsotsos, Iuliia Kotseruba\\
York University\\
{\tt\small {tsotsos, yulia\_k}@eecs.yorku.ca}
\and
Alexander Andreopoulos\\
IBM Research\\
{\tt\small aandreo@us.ibm.com}
\and
Yulong Wu\\
Aion Foundation\\
{\tt\small yulong@aion.network}
}

\maketitle
\ificcvfinal\thispagestyle{empty}\fi

\begin{abstract}
  Why do some continue to wonder about the success and dominance of deep learning methods in computer vision and AI? Is it not enough that these methods provide practical solutions to many problems? Well no, it is not enough, at least for those who feel there should be a science that underpins all of this and that we should have a clear understanding of how this success was achieved. Here, this paper proposes that the dominance we are witnessing would not have been possible by the methods of deep learning alone: the tacit change has been the evolution of empirical practice in computer vision and AI over the past decades. We demonstrate this by examining the distribution of sensor settings in vision datasets and performance of both classic
and deep learning algorithms under various camera settings. This reveals a strong mismatch between optimal performance ranges of classical theory-driven algorithms and sensor setting distributions in the common vision datasets, while data-driven models were trained for those datasets. The head-to-head comparisons between data-driven and theory-driven models were therefore unknowingly biased against the theory-driven models.
\end{abstract}

\section{Introduction}

There are many classic volumes that define the field of computer vision (e.g., Rosenfeld 1976 \cite{rosenfeld1976digital}, Marr 1982 \cite{marr1982vision}, Ballard \& Brown 1982 \cite{ballard1982computer}, Horn 1986 \cite{horn1986robot}, Koenderink 1990 \cite{koenderink1990solid}, Faugeras 1993 \cite{faugeras1993three}, Szeliski 2010 \cite{szeliski2010computer} and more). There, the theoretical foundations of image and video processing, analysis, and perception are developed both theoretically and practically. In part, these represent what we call here \textit{theory-driven computer vision models}.

A geometrical and physical understanding of the image formation process, from illuminant to camera optics to image creation, as well as the material properties of the surfaces that interact with incident light and the motions of objects in a scene, was mathematically modeled in the hope that, when those equations were simulated by a computer, they would reveal the kinds of structures that human vision also perceives. An excellent example of this development process appears in \cite{zucker1987vision}, which details the importance of physical and mathematical constraints for the theory development process with many examples from early work on edges, textures and shape analysis. It is difficult to deny the theoretical validity of those approaches and from the earliest days of computer vision, the performance of these theory solutions had always appeared promising, with a large literature supporting this (see the several articles in \cite{shapiro1992encyclopedia} on all aspects of computer vision for reviews of early work) as well as many commercial successes.

However, during most of the history of computer vision, the discipline suffered from two main problems (see \cite{andreopoulos201350}). Firstly, computational power and memory were too meagre to deal with the requirements of vision (theoretically shown in \cite{tsotsos1989complexity, tsotsos1987complexity}). Secondly, the availability of large sets of test data that could be shared and could permit replication of results was limited. An appropriate empirical methodology and tradition to guide such testing and replication was also missing. 

The first problem gradually improved as Moore's Law played out. Especially important, was the advent of GPUs in the late 1990s \cite{nickolls2010gpu}, with their general availability a few years later. Major progress was made on the second problem with the introduction of what might be the first large collection of images for computer vision testing, namely, MNIST \cite{lecun1998mnist}.  During the preceding decade, advances in machine learning began to have an important impact on computer vision (e.g., Turk \& Pentland Eigenface system \cite{turk1991face}, based on learning low-dimensional representations for faces, following Sirovich \& Kirby \cite{sirovich1987low}). 

Whereas the scarcity of data precluded extensive use of learning methods in the early days, the emergence of large image sets encouraged exploration of how systems might learn regularities, say in object appearance, over large sets of exemplars. Earlier papers pointed to the utility of images of handwritten digits in testing recognition and learning methods (e.g., work by Hinton et al. \cite{hinton1992adaptive}) so the creation of the MNIST set was both timely and impactful. As a result, the community witnessed the emergence of \textit{data-driven computer vision models} created by extracting statistical regularities from a large number of image samples. The MNIST set was soon joined by others. The PASCAL Visual Object Classes (VOC) Challenge began in 2005 \cite{everingham20052005}, ImageNet in 2010 \cite{russakovsky2015imagenet}, and more. The contribution of these data sets and challenges is undeniable towards the acceleration of developments in computer vision.

It is widely accepted that the AlexNet paper \cite{krizhevsky2012imagenet} was a turning point in the discipline. It outperformed theory-driven approaches by a wide margin and thus began the surrender of those methods to data-driven ones. But how did that success come about? Was it simply the fact that AlexNet was a great feat of engineering? Likely this is true; but here, we suggest that there may be a bit more to it. Empirical methodology matters. How this has evolved within the community had also played a role, perhaps a key role, so that the theory-driven computer vision paradigms never had a chance. In the rush to celebrate and exploit this success, no one noticed. 

The next sections will play out our path to supporting this assertion. It is interesting to note that what led us to this was work towards understanding how active setting of camera parameters affects certain computer vision algorithms. We summarize the argument here.

\section{Effect of Sensor Settings for Interest Point and Saliency Algorithms}

Previous work explored the use of interest point/saliency algorithms in an active sensing context and investigated how they perform with varying camera parameters in order to develop a method for dynamically setting those parameters depending on task and context. The experiments revealed a very strong dependence on settings for a range of algorithms. The patterns seemed orderly as if determined by some physical law; the data exhibited a strong and clear structure. A brief summary is provided here with more detail in Andreopoulos \& Tsotsos \cite{andreopoulos2011sensor}.

The authors evaluated the effects of camera shutter speed and voltage gain, under simultaneous changes in illumination, and demonstrated significant differences in the sensitivities of popular vision algorithms under those variations. They began by creating a dataset that reflected different cameras, camera settings, and illumination levels. They used one CCD sensor (Bumblebee stereo camera, PointGrey Research) and one CMOS sensor (FireflyMV camera, PointGrey Research) to obtain the datasets. The permissible shutter exposure time and gain were uniformly quantized with 8 samples in each dimension. There were an equal number of samples for each combination of sensor settings.

\begin{figure}[ht!]
\centering
\begin{subfigure}[b]{0.48\columnwidth}
\includegraphics[width=\columnwidth]{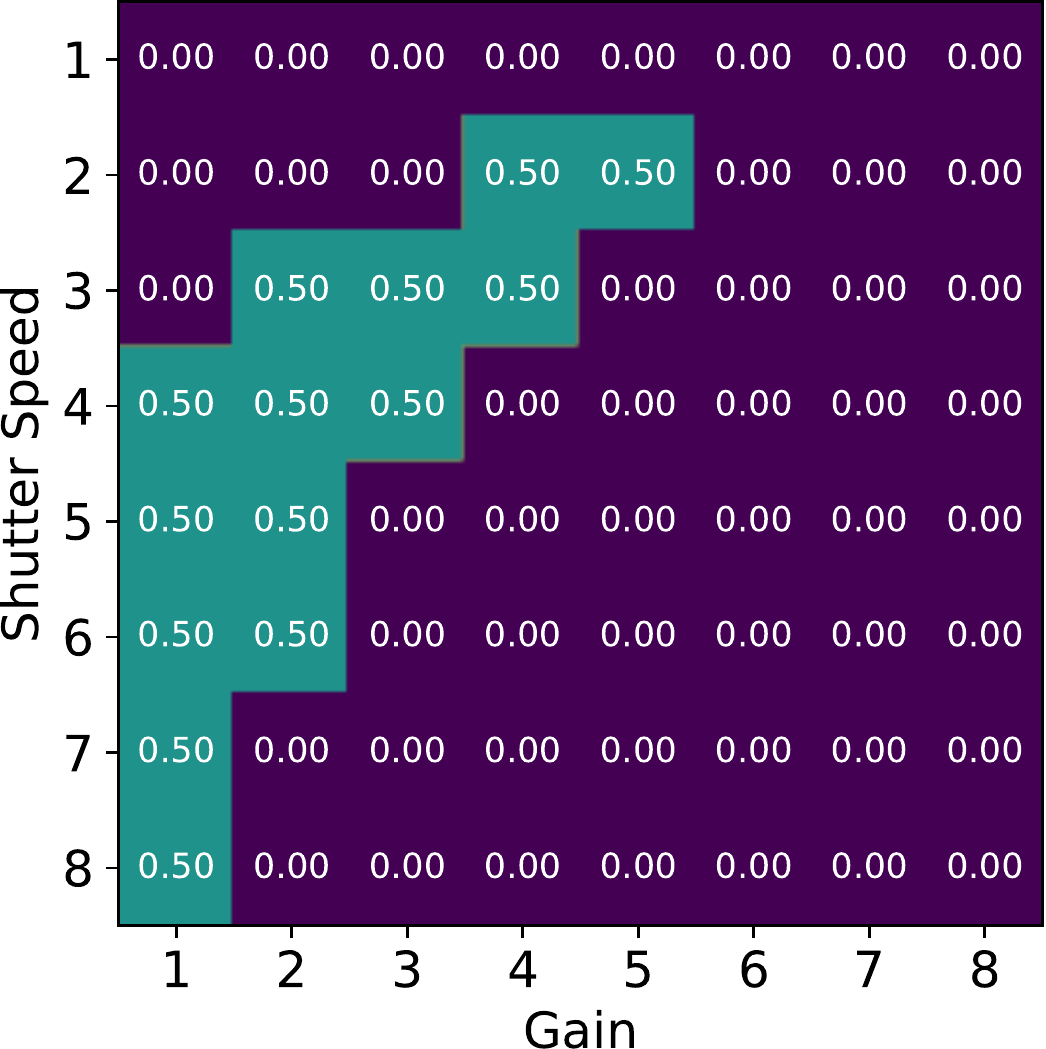}
\caption{\label{Harris}Harris}
\end{subfigure}
\begin{subfigure}[b]{0.48\columnwidth}
\includegraphics[width=\columnwidth]{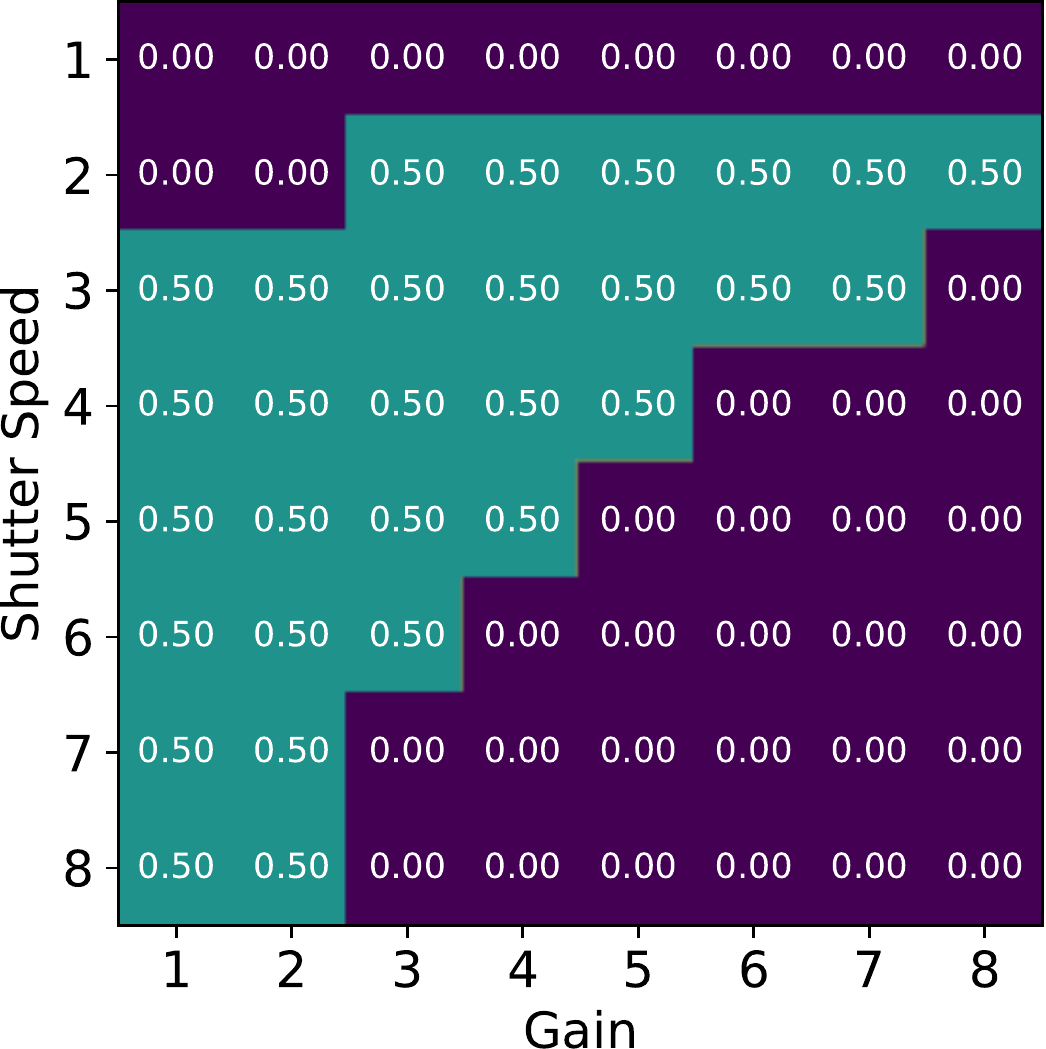}
\caption{\label{Hessian}Hessian}
\end{subfigure}
\begin{subfigure}[b]{0.48\columnwidth}
\includegraphics[width=\columnwidth]{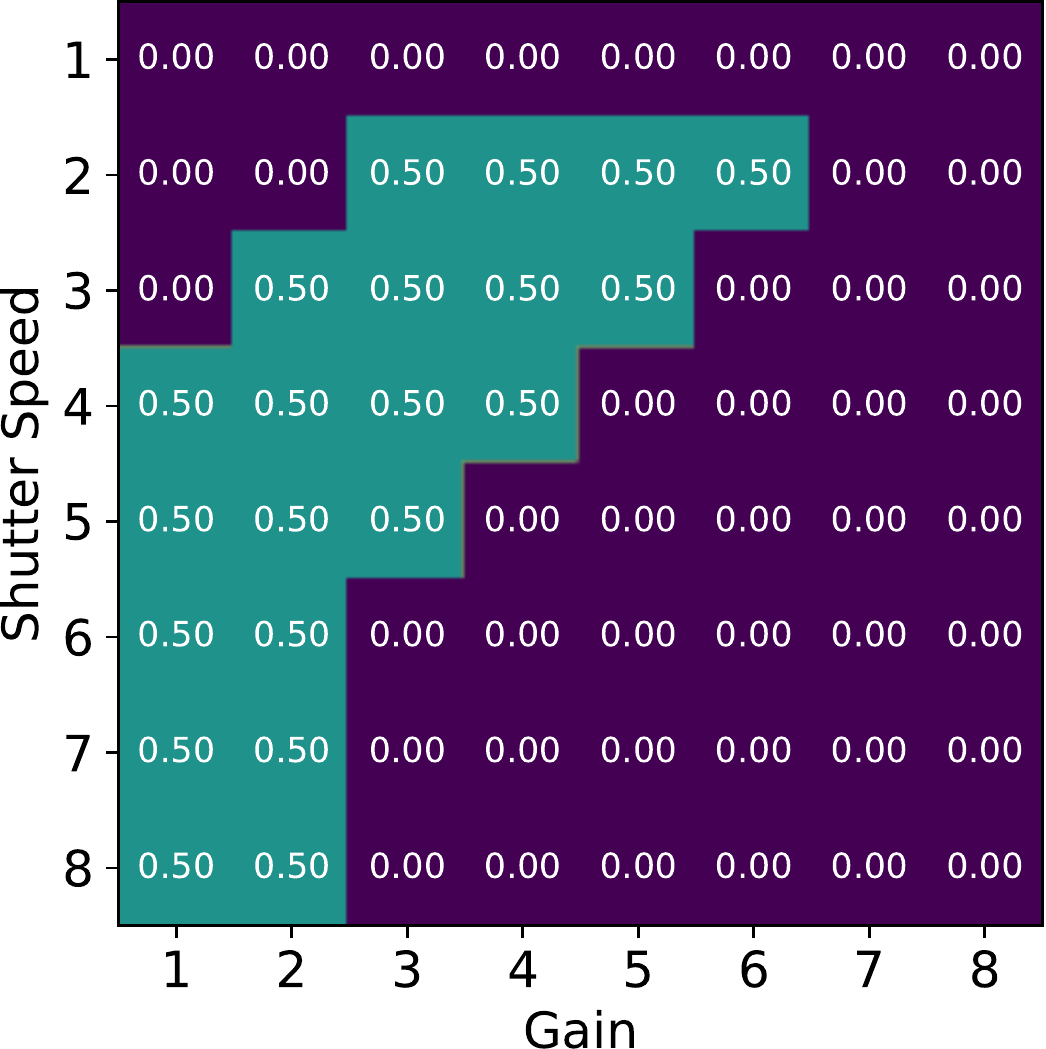}
\caption{\label{MSER}MSER}
\end{subfigure}
\begin{subfigure}[b]{0.48\columnwidth}
\includegraphics[width=\columnwidth]{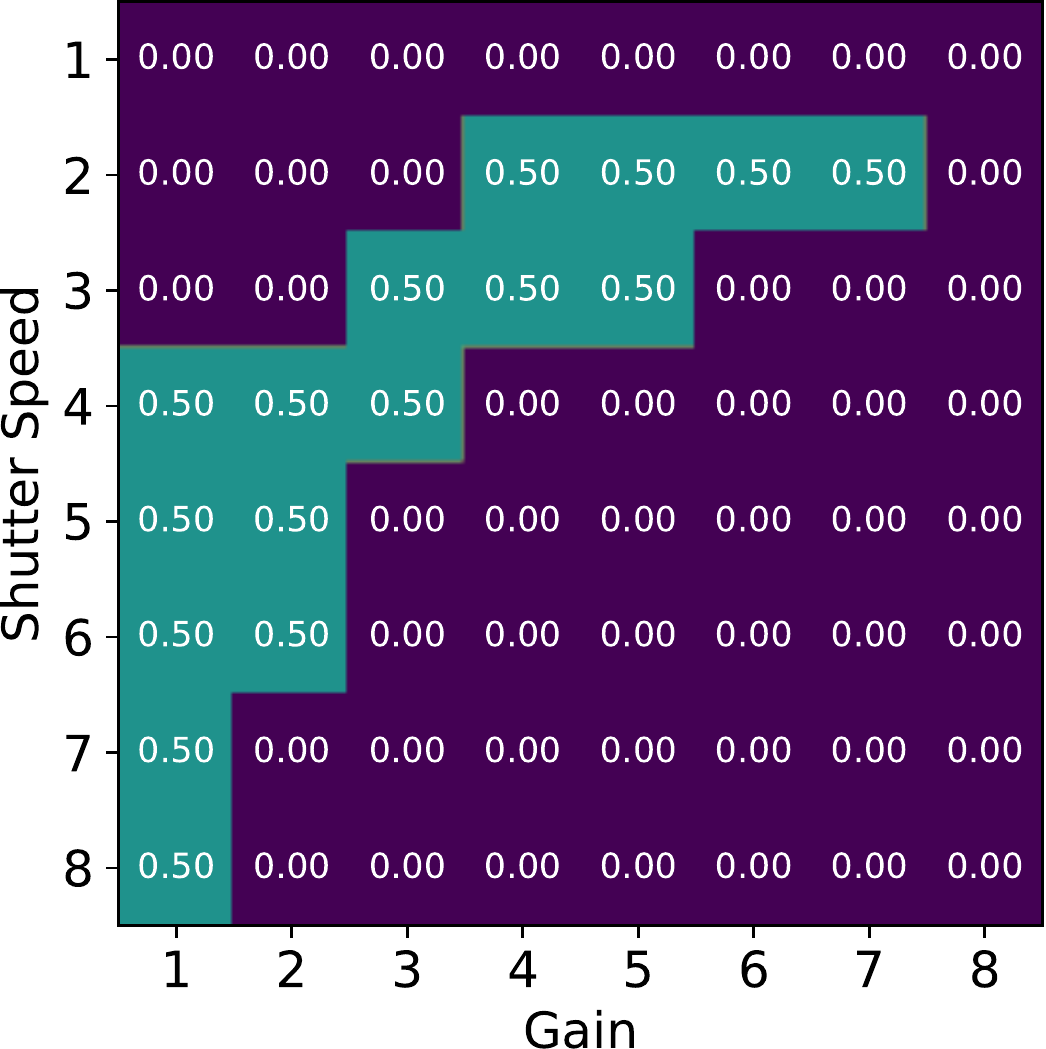}
\caption{\label{SURF}SURF}
\end{subfigure}
\begin{subfigure}[b]{0.48\columnwidth}
\includegraphics[width=\columnwidth]{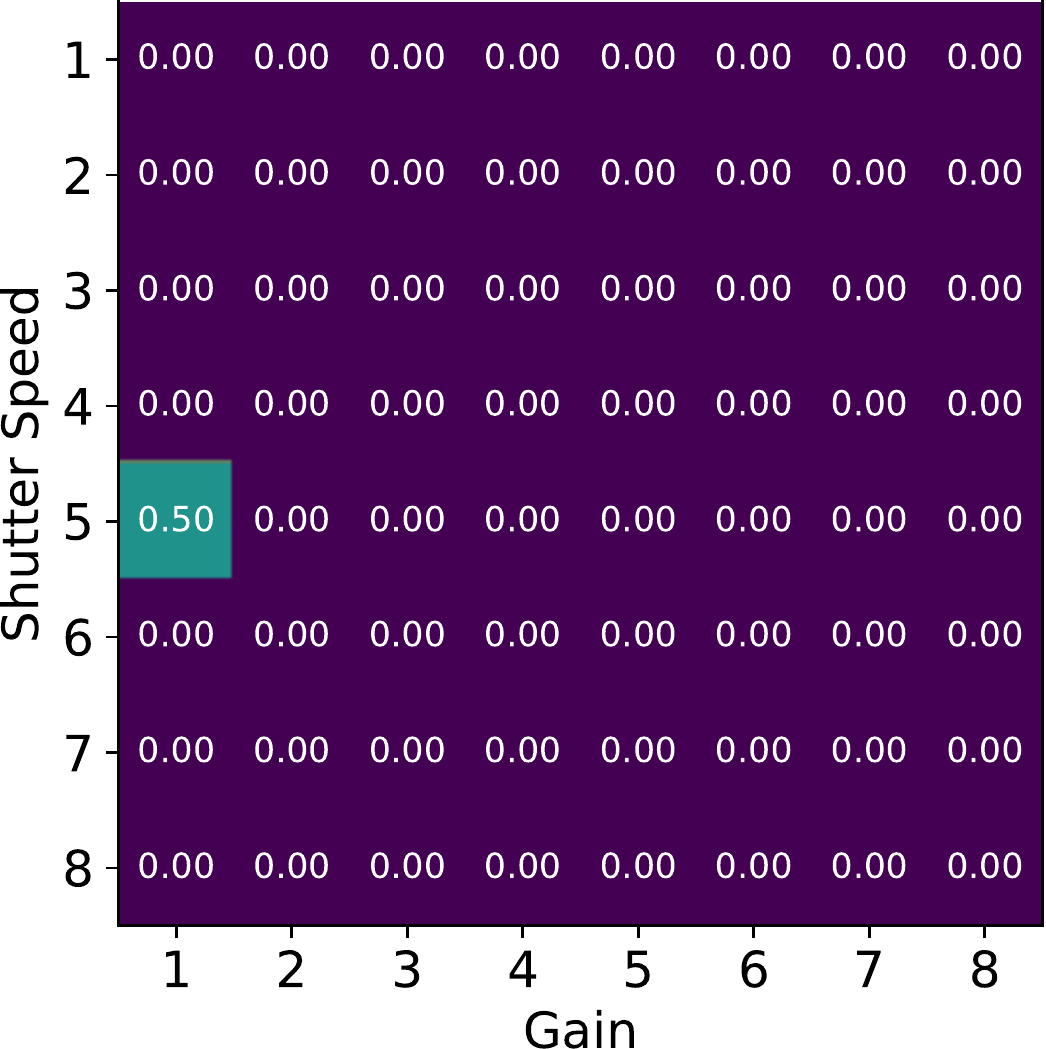}
\caption{\label{ScaleSaliency}Scale Saliency}
\end{subfigure}
\caption{\label{alex_eval} Adapted from Andreopoulos \& Tsotsos \cite{andreopoulos2011sensor} showing the performance of various descriptors in terms of precision-recall values for each combination of sensor settings (collapsed across all illumination conditions tested). The descriptors tested are (a) Harris-Affine; (b) Hessian-Affine; (c) MSER; (d) SURF; (e) Scale Saliency. In all plots shutter speed increases from top to bottom and gain increases from left to right. In \cite{andreopoulos2011sensor} precision and recall values are thresholded at 0.5, so we set the appropriate bins to 0.5 in the images above.}
\vspace{-0.5em}
\end{figure}

Then, the resulting images were processed by several algorithms, popular before the time of writing:  Harris-Affine and Hessian-Affine region detectors  \cite{mikolajczyk2004scale}, Scale Saliency interest-point detector \cite{kadir2001saliency}, Maximally Stable Extremal Regions algorithm (MSER) \cite{matas2004robust}, and Speeded-Up Robust Features extraction algorithm (SURF) \cite{bay2006surf}. 

Performance was determined by first selecting a target image for each scenario (camera setting and illumination combination) which
became the image with respect to which the change in the detected features under different shutter/gain values and different scene illumination is evaluated for a different image. The target image is identical regardless of the interest-point or saliency algorithm being tested for a given dataset, so that the results acquired with different algorithms are comparable (the scene is unchanging). Precision and recall are then determined based on how well an algorithm detects those target features as the scenario varies for the other images in the dataset. This is fully detailed in \cite{andreopoulos2011sensor}.

 The results, summarized in Figure \ref{alex_eval}, show a strong structure defining the good performance ranges of each algorithm. They all are have different shape and point out that sensor and illumination settings that lead to good performance are particular to each algorithm. This suggests that common datasets used to evaluate vision algorithms may suffer from a significant sensor specific bias which can make many of the experimental methodologies used to evaluate vision algorithms unable to provide results that generalize in less controlled environments. Simply put, if one wished to use one of these specific algorithms for a particular
application, then it is necessary to ensure that the images processed are acquired using the sensor setting ranges that yield good performance (see Figure 1). Such considerations are rarely observed.

Also tested in \cite{andreopoulos2011sensor} but not shown here are the saliency algorithms of Itti-Koch-Niebur \cite{itti1998model} and the Bruce-Tsotsos AIM algorithm \cite{bruce2009saliency}). Those tests also revealed that good performance depends on different illumination conditions for different sensor settings. This demonstrates the inability of a single constant shutter/gain value to optimally discern the most salient image regions under modest changes in the illumination, or as the surface reflectance properties change. Some of their worst performance appeared when using the auto-gain and auto-exposure mechanism.  Such sensor mechanisms, which rely on the mean image intensity, are very commonly used in practice yet lead to extremely poor and inconsistent results when those images are processed using any of these algorithms.

\section{Effect of Sensor Settings on Object Detection Algorithms}

The same test for more recent recognition algorithms, both classic and deep learning methods, was performed by Wu \& Tsotsos \cite{wu2017active}. The authors created a dataset containing 2240 images in total, by viewing 5 different objects (bicycle, bottle, chair, potted plant and tv monitor), at 7 levels of illumination and with 64 camera configurations (8 shutter speeds, 8 voltage gains). As in the previous experiment, there were an equal number of samples for each combination of sensor settings. To accurately measure the illumination of the scene, a Yoctopuce light sensor was used. Also, intensity-controllable light bulbs were used to achieve different light conditions, 50lx, 200lx, 400lx, 800lx, 1600lx and 3200lx. The digital camera was a Point Grey Flea3. The allowed ranges of shutter speed and voltage gain were uniformly sampled into 8 distinct values in each dimension.

\begin{figure}[ht!]
\centering
\begin{subfigure}[b]{0.48\columnwidth}
\includegraphics[width=\columnwidth]{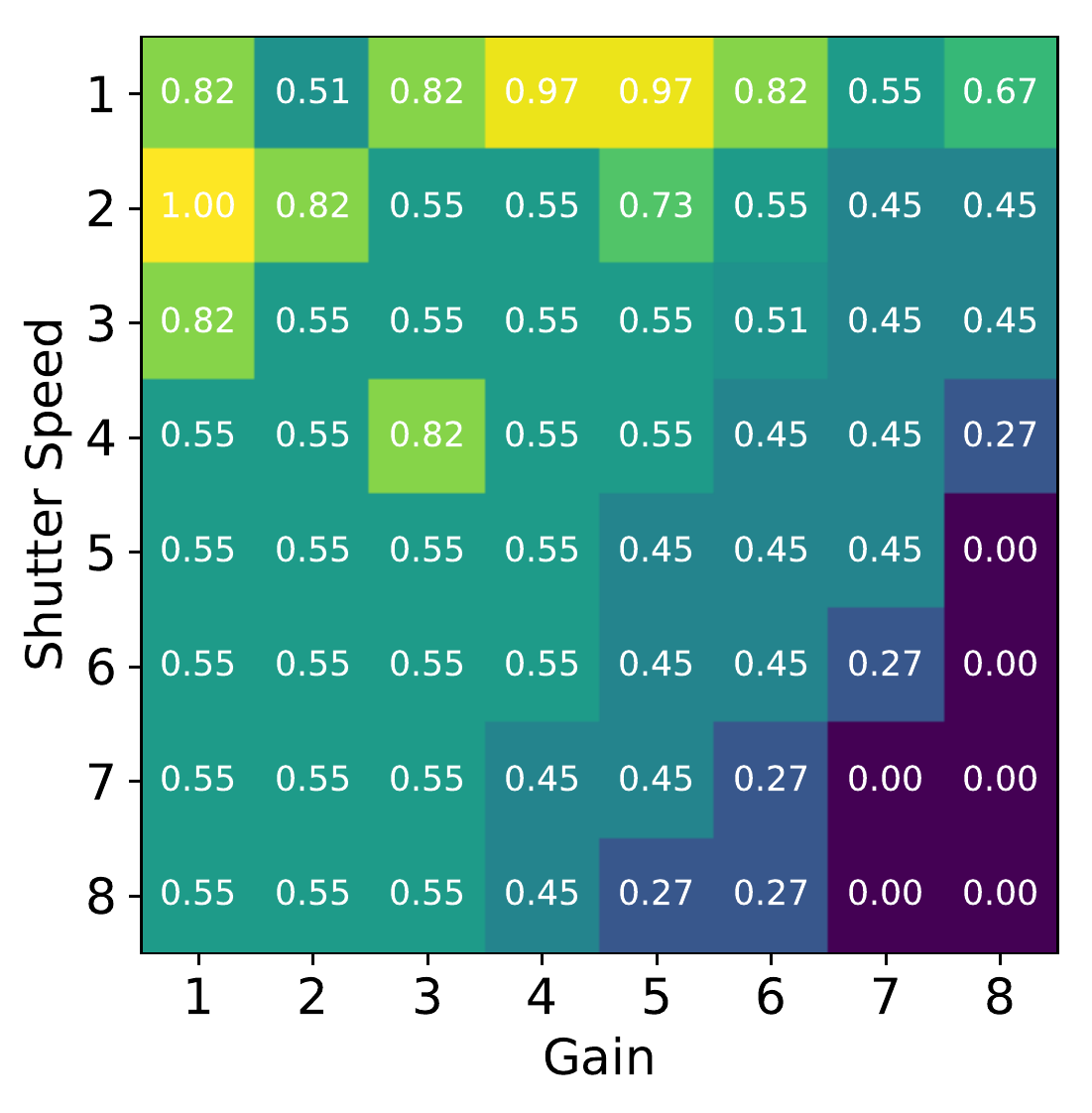}
\caption{\label{RCNN}R-CNN}
\end{subfigure}
\begin{subfigure}[b]{0.48\columnwidth}
\includegraphics[width=\columnwidth]{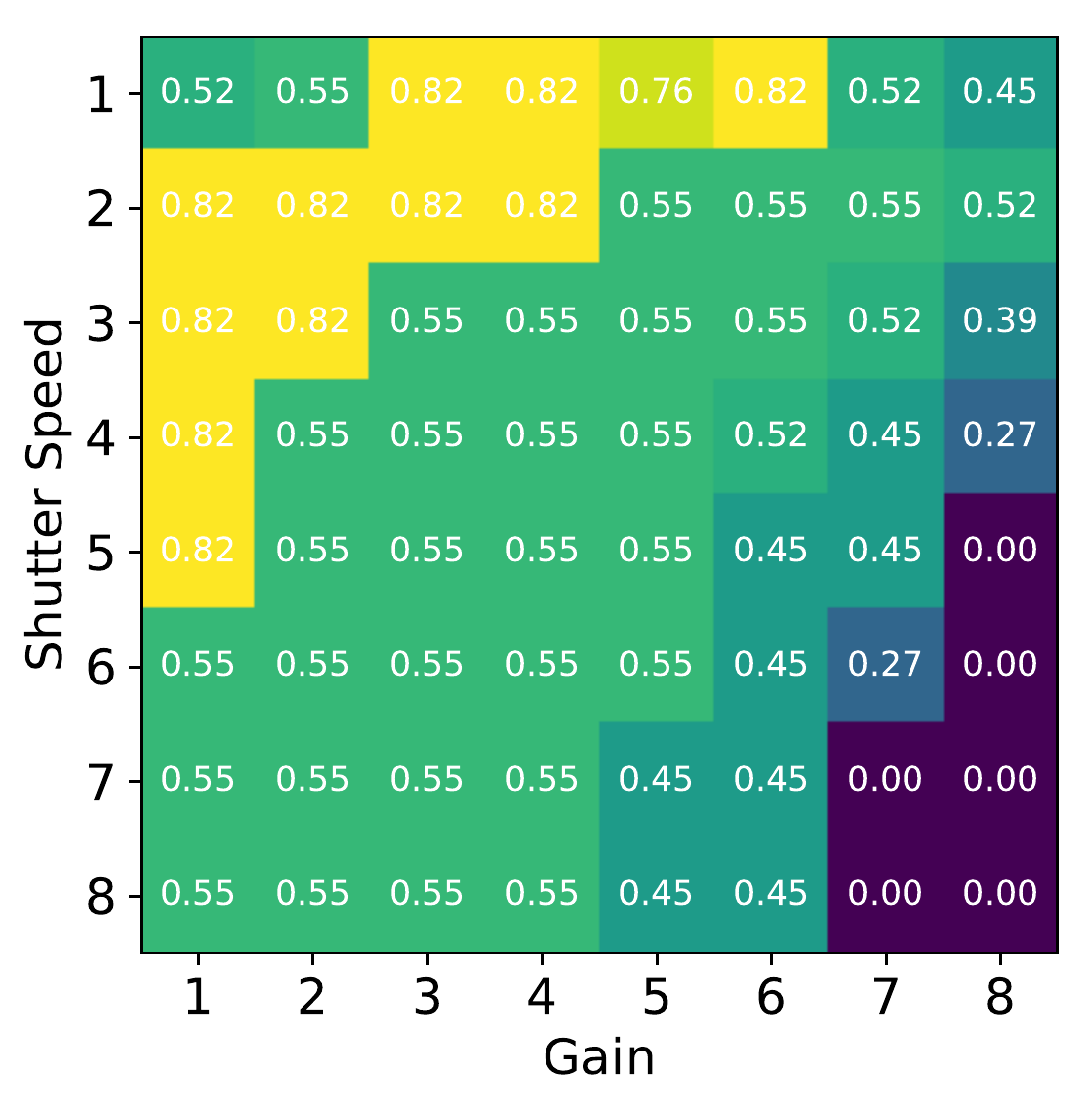}
\caption{\label{SPPNet}SPP-net}
\end{subfigure}
\begin{subfigure}[b]{0.48\columnwidth}
\includegraphics[width=\columnwidth]{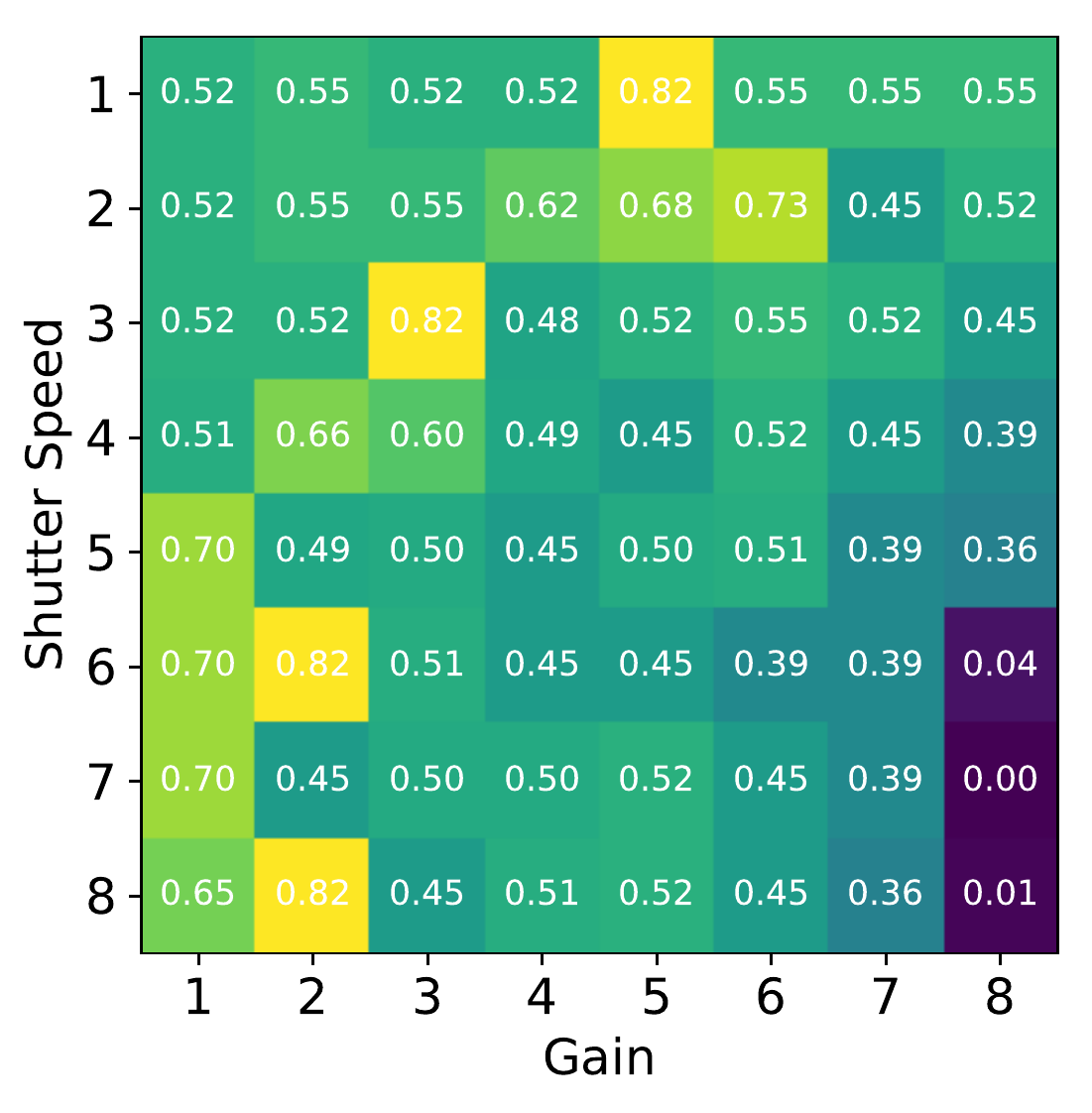}
\caption{\label{DPM}DPM}
\end{subfigure}
\begin{subfigure}[b]{0.48\columnwidth}
\includegraphics[width=\columnwidth]{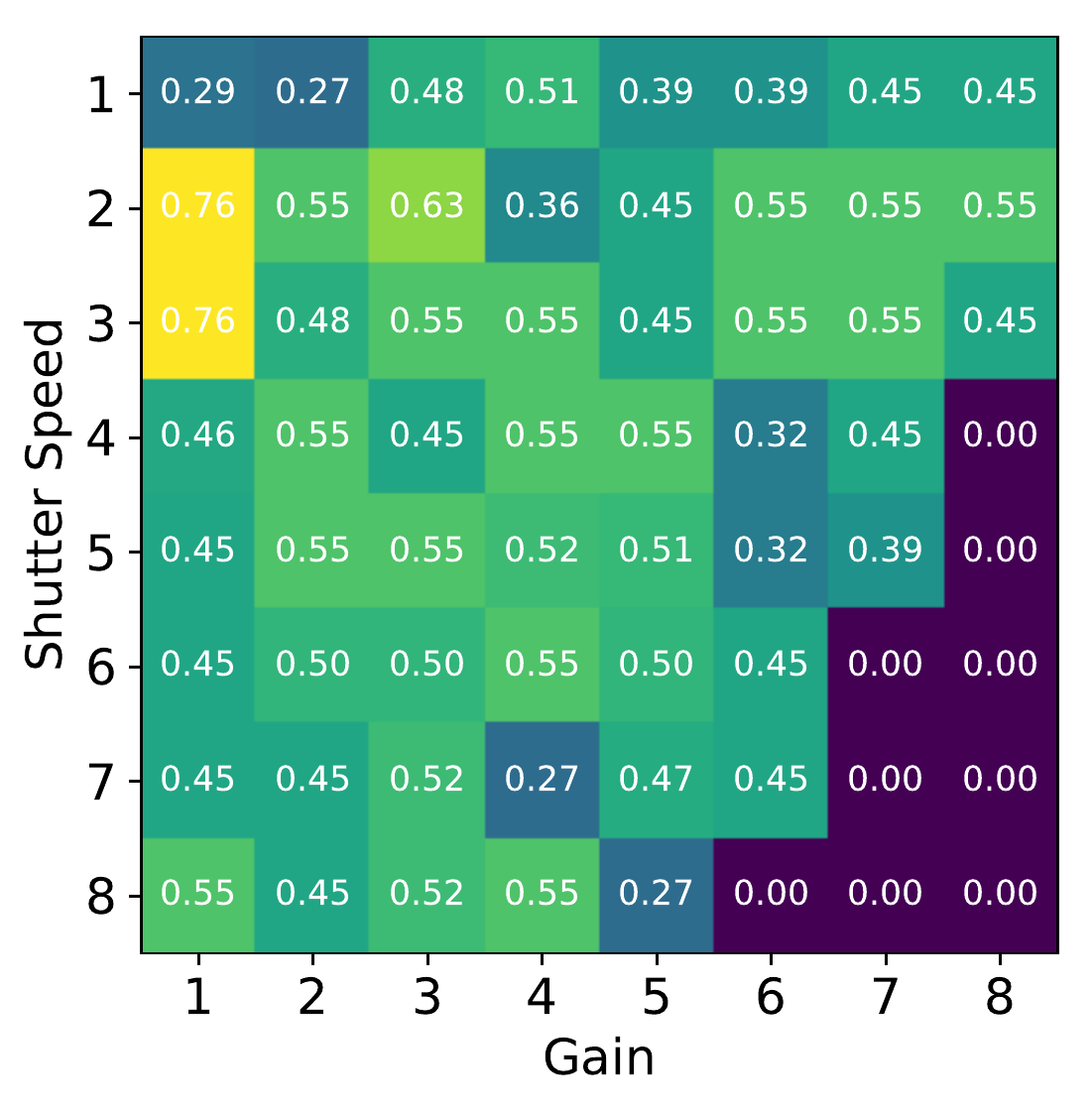}
\caption{\label{Bow}BoW}
\end{subfigure}
\caption{\label{wu_eval} Results of evaluation of object detection algorithms on the custom dataset for different sets of shutter speed and gain values adapted from Wu \& Tsotsos \cite{wu2017active}. In all plots shutter speed increases from top to bottom and gain increases from left to right. As in Figure \ref{alex_eval}, mAP values (as defined in \cite{salton1983introduction}) are shown for each sensor setting combination and a particular illumination level. Here we show the results for the high illumination conditions, which most closely reflect what is found in the common vision datasets.}
\vspace{-0.5em}
\end{figure}

Four popular object detection algorithms were selected for evaluation: the Deformable Part Models (DPM) \cite{felzenszwalb2009object}, the Bag-of-Words Model with Spatial Pyramid Matching (BoW) \cite{uijlings2013selective}, the Regions with Convolutional Neural Networks (R-CNN) \cite{girshick2014rich}, and the Spatial Pyramid Pooling in Deep Convolutional Networks (SPP-net) \cite{he2015spatial}. 

The results are shown in Figure \ref{wu_eval}. This time, mean average precision (mAP) values were not thresholded and are plotted intact. The original work presented these plots for each of the tested illumination levels and demonstrated that performance depends significantly on illumination level as well as sensor settings and does not easily generalize across these variables. As before, if one wished to use one of these specific algorithms for a particular application, then it is necessary to ensure that the images processed are acquired using the sensor setting ranges that yield good performance (see Figure \ref{wu_eval}). 

In general, it can be seen that there is less orderly structure when compared to the previous set of tests (thus making any characterization of `good performing' sensor settings more difficult) and the authors wondered about the reason underlying this difference. Could the difference be due to an uneven distribution of training samples along those dimensions? And if so, could overall performance be influenced by such bias?

\section{Distributions of Sensor Parameters in Common Computer Vision Datasets}

As mentioned, the two above studies caused us to be curious about the reasons behind the uneven and unexpected performance patterns across algorithms.  After thorough verifications of the methods employed, we concluded that some imbalance in data distribution across sensing parameters might be the cause. 

Surprisingly, among works on various biases in vision datasets, few acknowledge the existence of sensor bias (or \textit{capture bias} \cite{tommasi2017deeper}) and none provide quantitative statistics. Hence, we set out to test this. We selected two common datasets, Common Objects in Context (COCO) \cite{lin2014microsoft} and VOC 2007, the dataset used in the PASCAL Visual Object Classes Challenge in 2007 \cite{pascal-voc-2007}.

Since both datasets consist of images gathered from Flickr, we used Flickr API to recover EXIF data (a set of tags for camera settings provided by the camera vendor) for each image. In the COCO dataset 59\% and 58\% of train and validation data respectively had EXIF data available. We use the \texttt{trainval35K} split commonly used for training object detection algorithms (e.g. SSD\cite{liu2016ssd}, YOLOv3 \cite{redmon2018yolov3}, and RetinaNet \cite{lin2017focal}).

\begin{figure}[ht!] 
\centering
\begin{subfigure}[b]{0.48\columnwidth}
\includegraphics[width=\columnwidth]{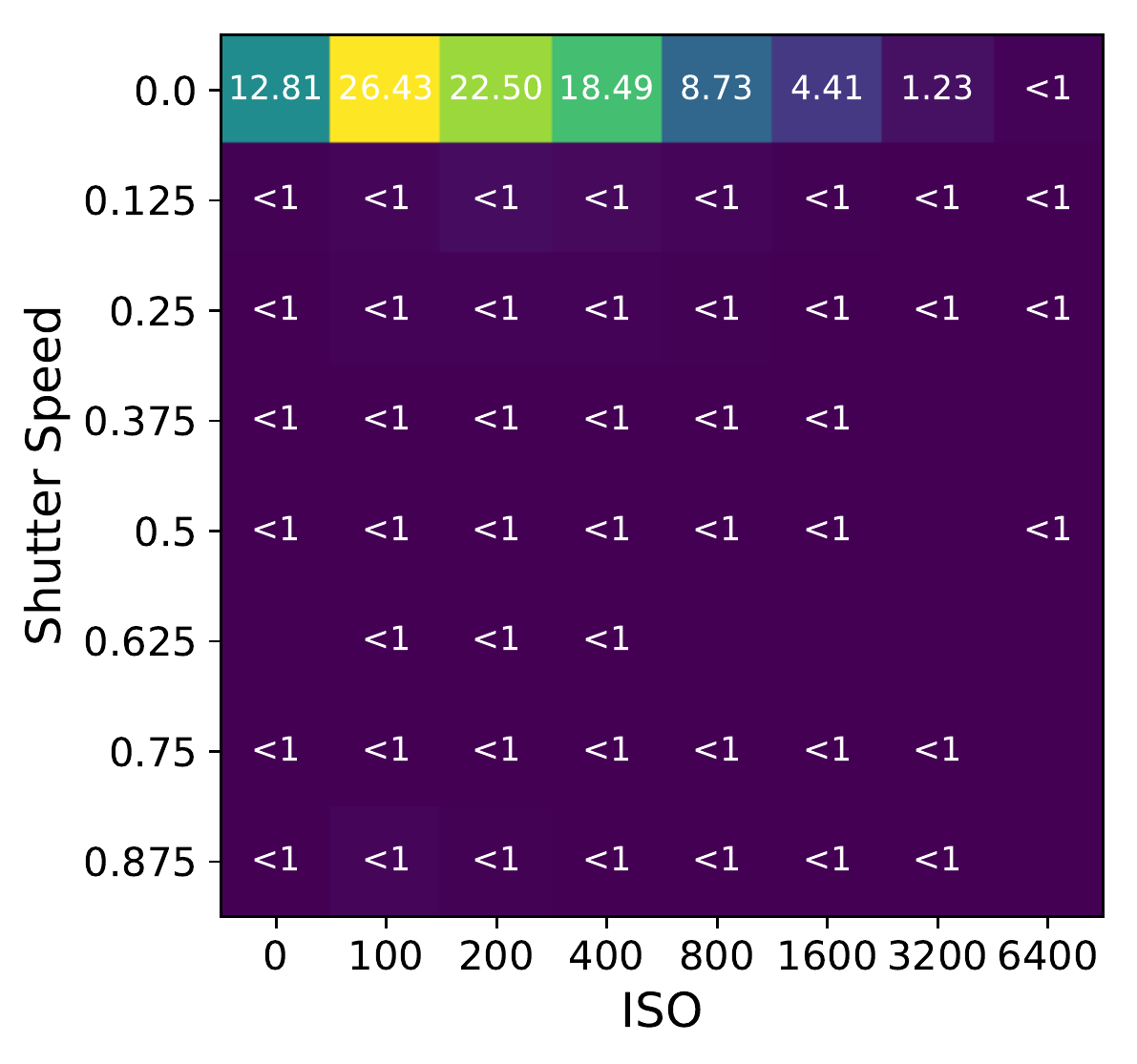}
\caption{\label{COCO_all_train} COCO train set}
\end{subfigure}
\begin{subfigure}[b]{0.48\columnwidth}
\includegraphics[width=\columnwidth]{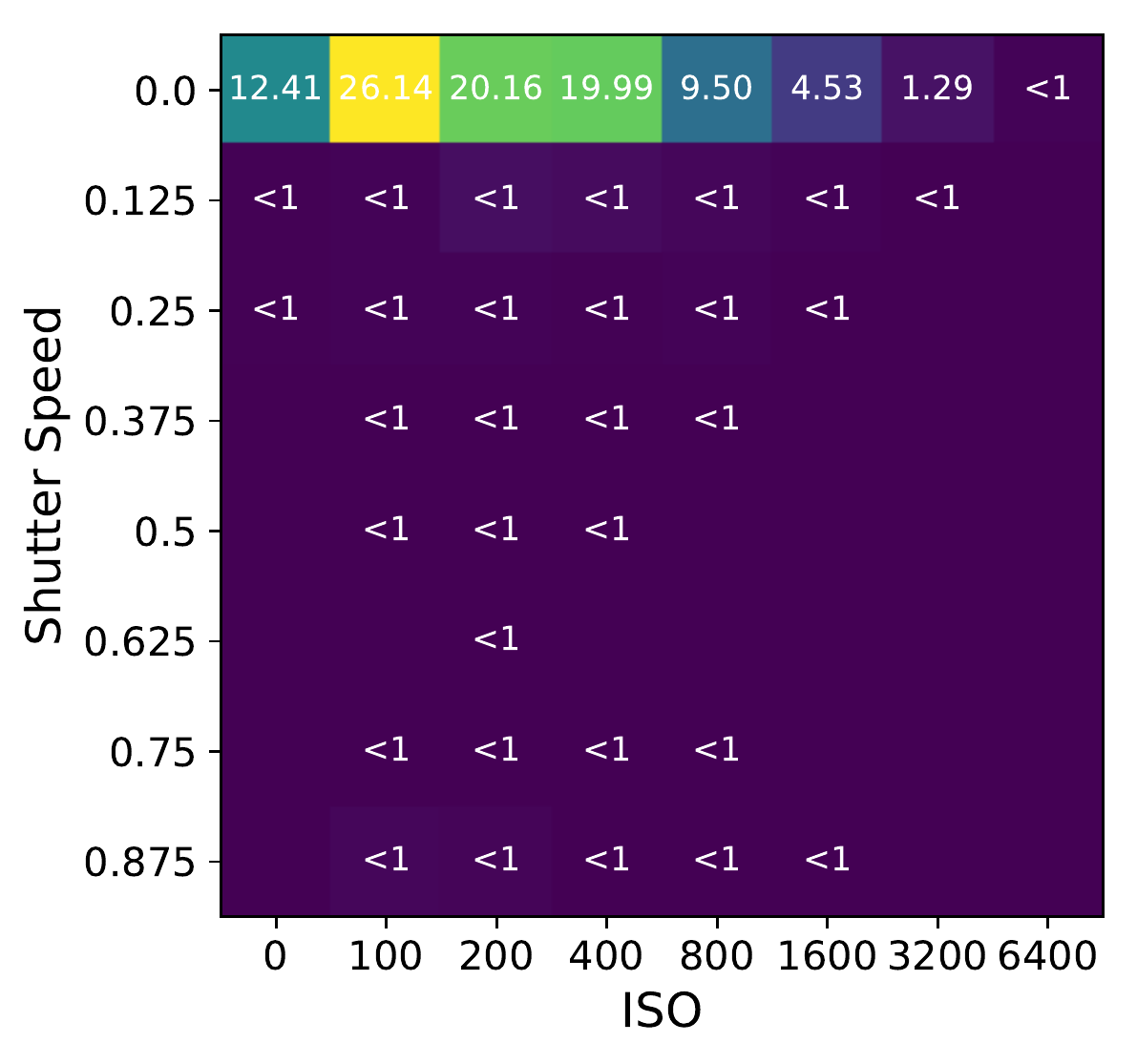}
\caption{\label{COCO_all_val} COCO val set}
\end{subfigure}
\begin{subfigure}[b]{0.48\columnwidth}
\includegraphics[width=\columnwidth]{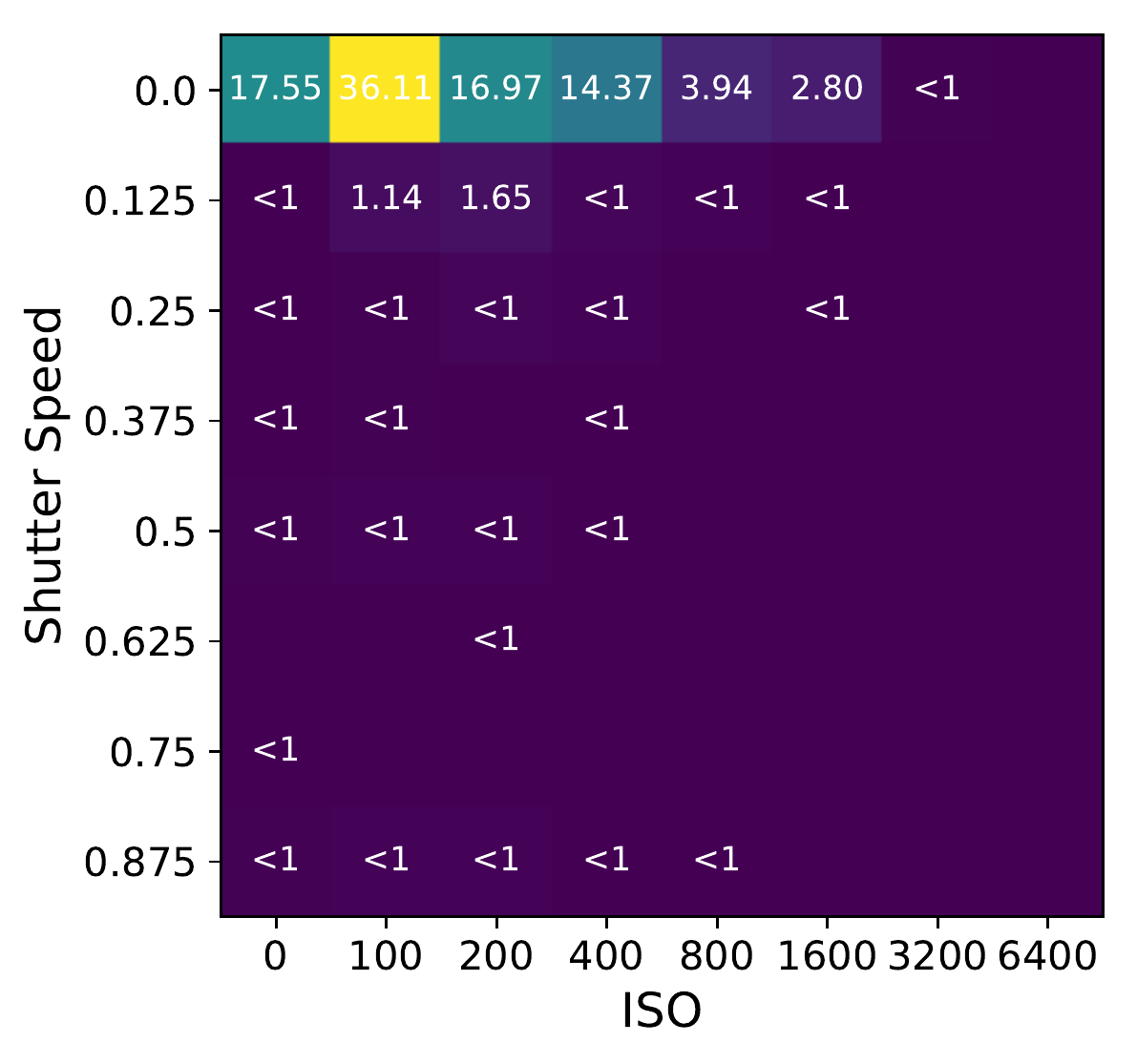}
\caption{\label{VOC2007_all_train} VOC2007 train set}
\end{subfigure}
\begin{subfigure}[b]{0.48\columnwidth}
\includegraphics[width=\columnwidth]{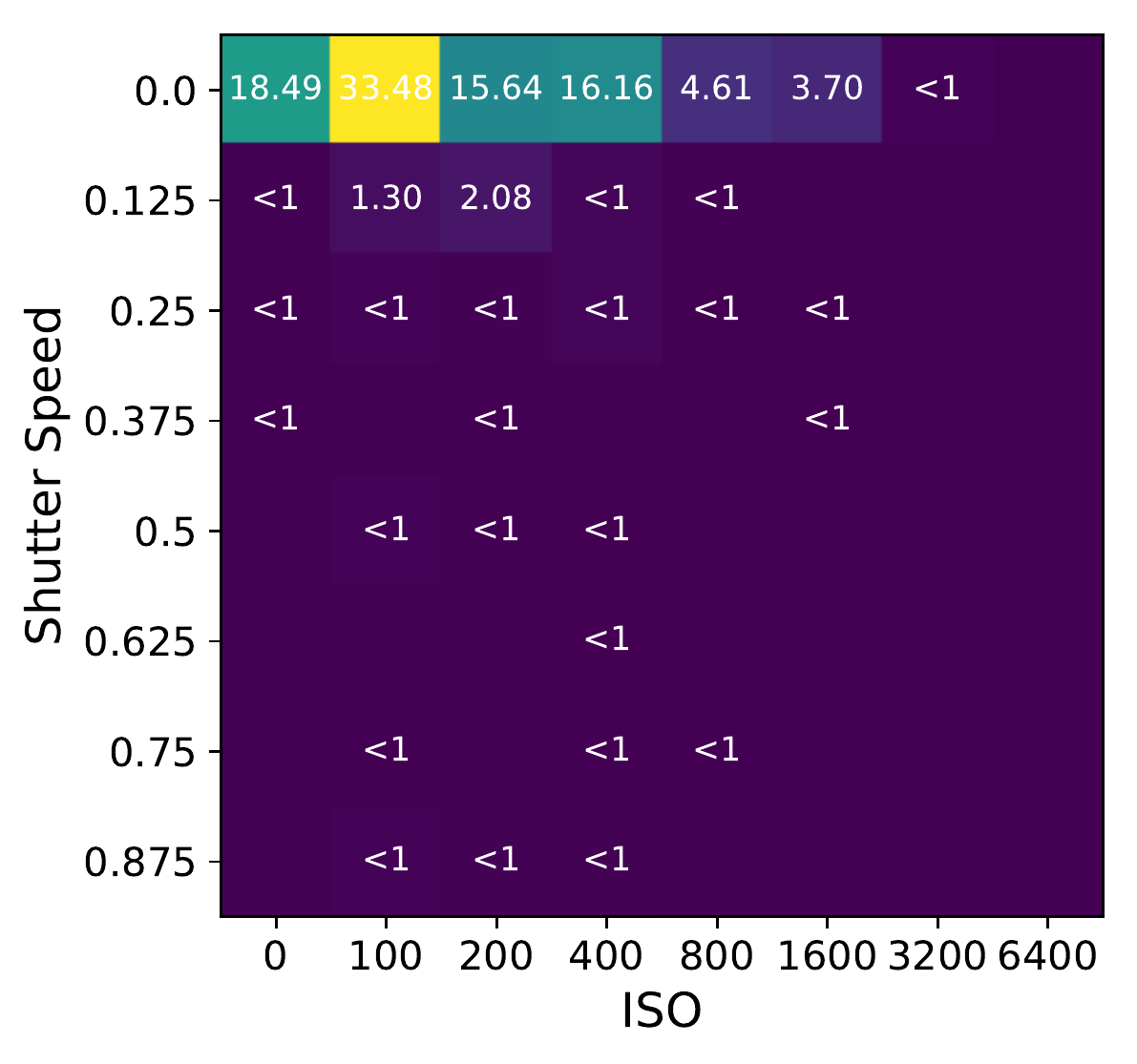}
\caption{\label{VOC2007_all_val} VOC2007 test set}
\end{subfigure}

\caption{\label{dataset_sensor_bias} Distribution of exposure time and ISO in training (a)  and validation (b) sets in terms of \% of the total images in the COCO set (with EXIF data). Distribution of exposure time and ISO in training (c) and test (d) sets in VOC2007 dataset (with available EXIF data). Note the uneven distribution of images in the bins. Some of the bins are empty since there are no images in the dataset obtained with those camera settings.}
\vspace{-0.5em}
\end{figure}  

\begin{figure}[ht!] 
\centering
\begin{subfigure}[b]{0.48\columnwidth}
\includegraphics[width=\columnwidth]{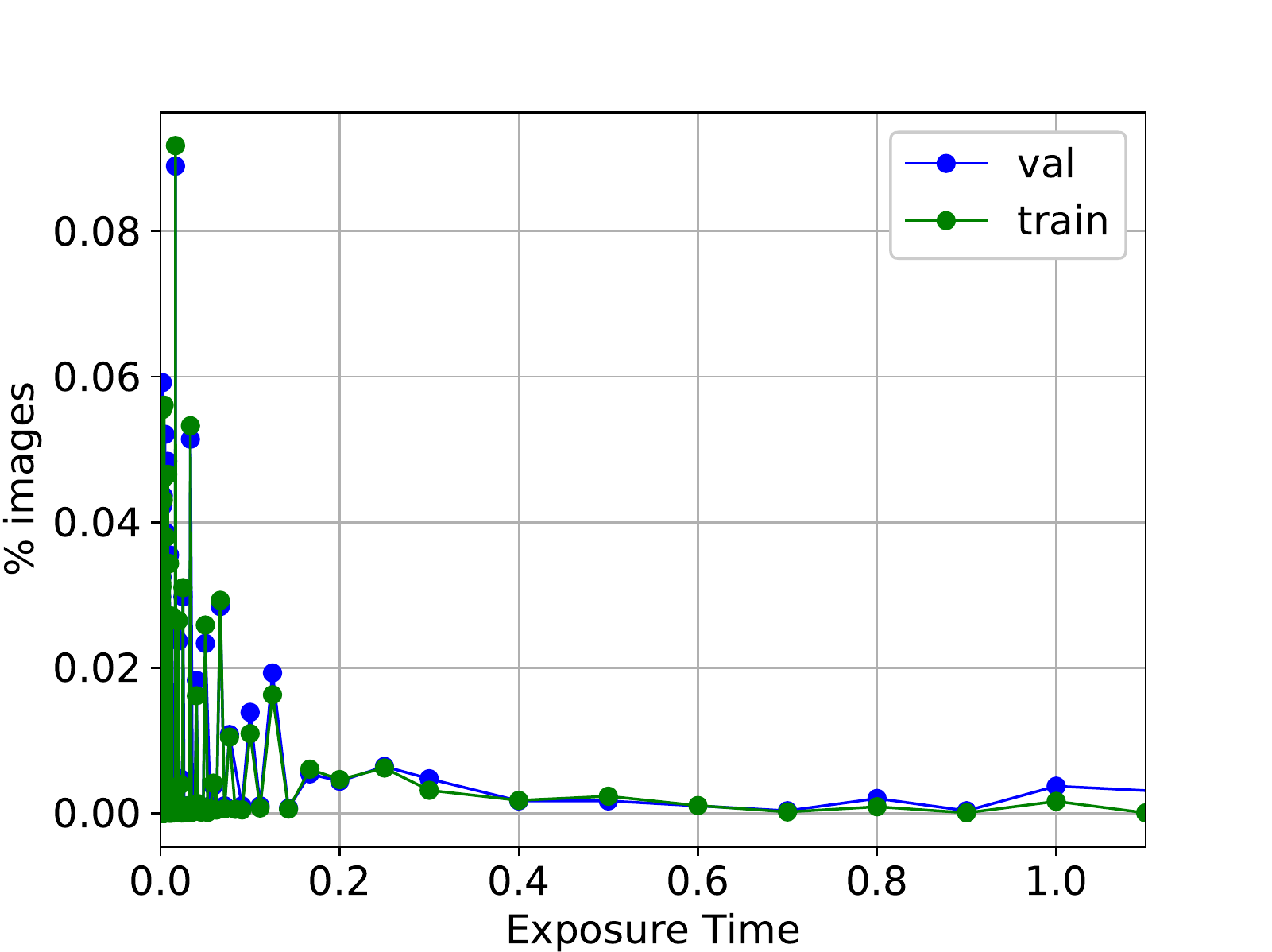}
\caption{\label{COCO_exp_distribution}{\footnotesize Exposures times in COCO}}
\end{subfigure}
\begin{subfigure}[b]{0.48\columnwidth}
\includegraphics[width=\columnwidth]{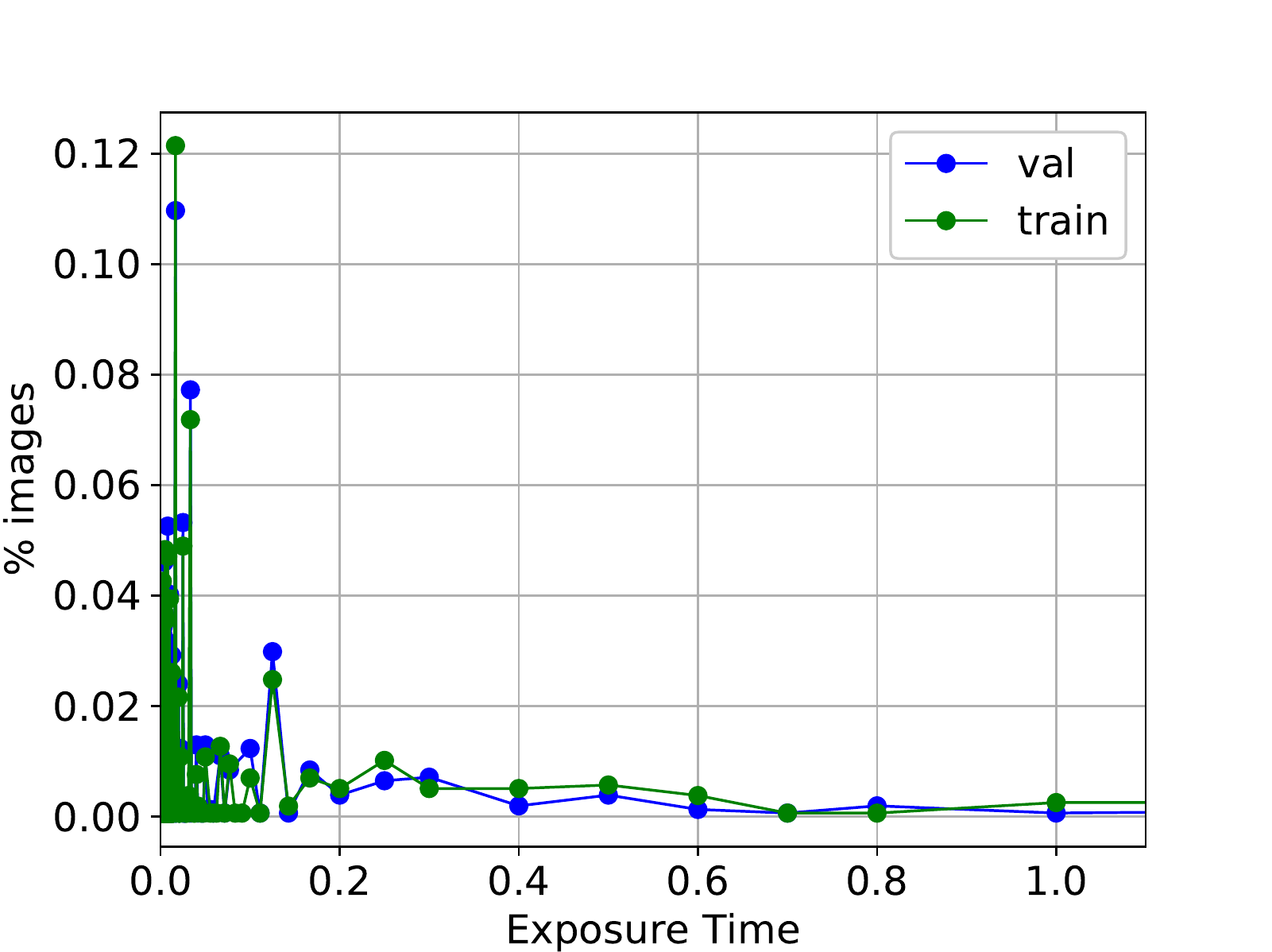}
\caption{\label{VOC_exp_distribution}{\footnotesize Exposure times in VOC2007}}
\end{subfigure}
\begin{subfigure}[b]{0.48\columnwidth}
\includegraphics[width=\columnwidth]{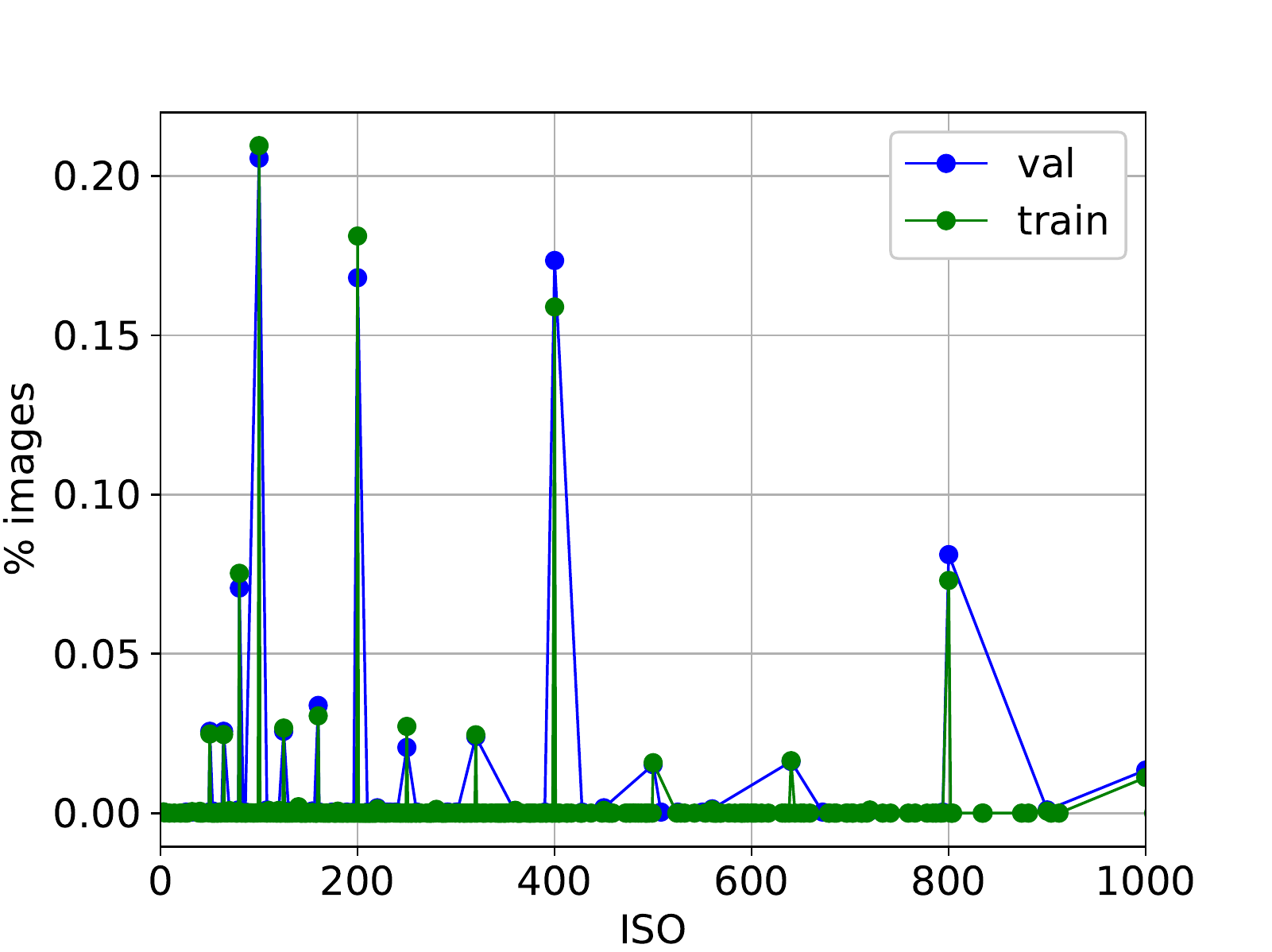}
\caption{\label{COCO_ISO_distribution}{\footnotesize ISO values in COCO}}
\end{subfigure}
\begin{subfigure}[b]{0.48\columnwidth}
\includegraphics[width=\columnwidth]{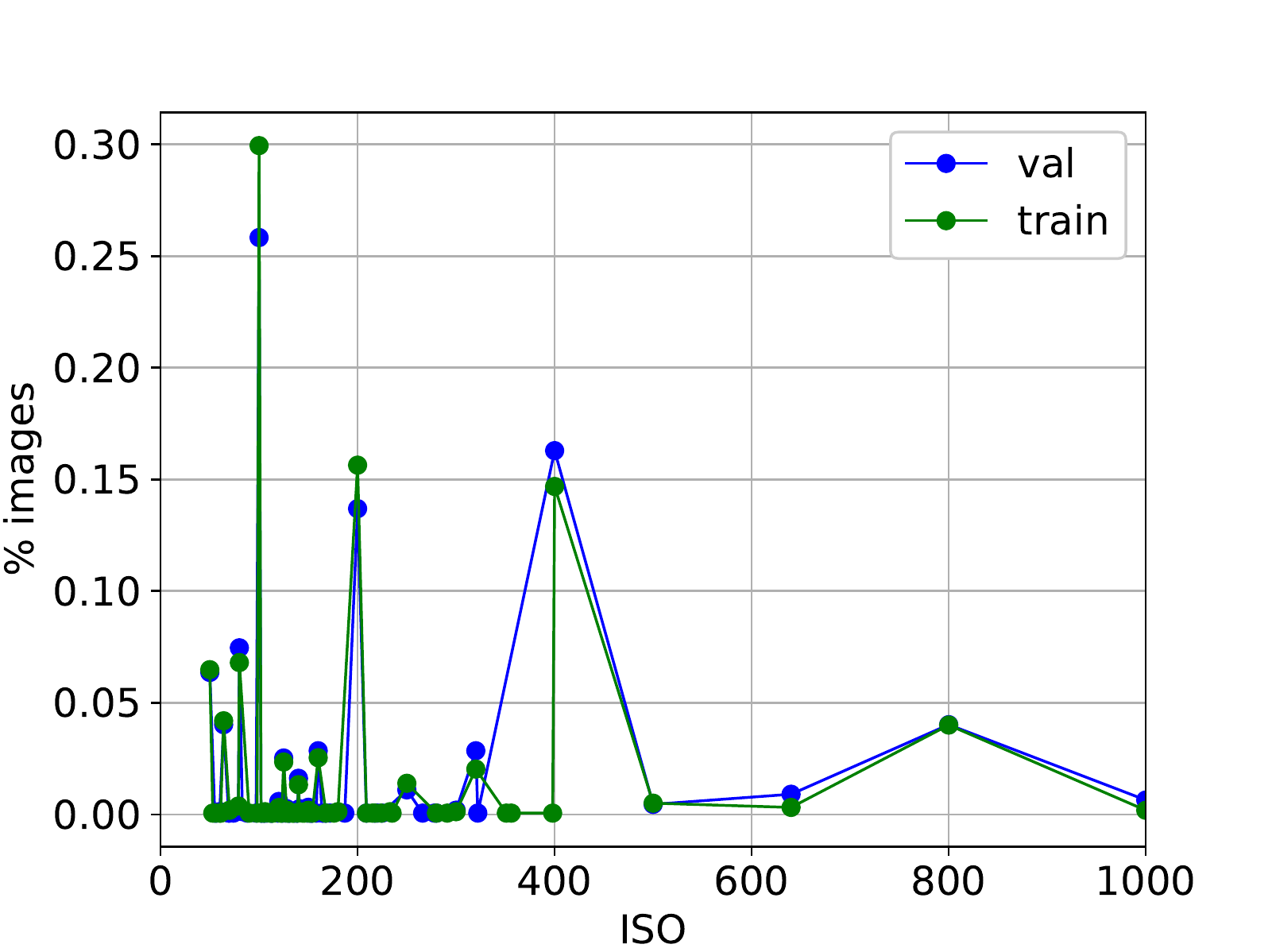}
\caption{\label{VOC_ISO_distribution}{\footnotesize ISO values in VOC2007}}
\end{subfigure}
\caption{\label{shutter_ISO_distributions} Distributions of exposure times (shutter speeds) in (a) COCO and (b) VOC2007 datasets. Distributions of ISO values in (c) COCO and (d) VOC2007 datasets. Train and validation data splits are shown in green and blue.}
\end{figure}

In the VOC2007 dataset  31\% of train and test data had EXIF data available, however as Figure \ref{dataset_sensor_bias} shows, the distributions of exposure times and ISO settings for photographs in the COCO and VOC2007 datasets are similar. Note that most images are taken with the automatic ISO settings such as 100, 200, 400 and 800 (recall the earlier test results regarding automatic camera settings). Most exposure times are short (below 0.1s) and spike around 1/60s. This agrees with the large-scale statistical analysis of millions of images found online in \cite{wueller2018statistic}.

Using shutter speed, f-number and ISO we can compute exposure value (EV) using the following formula as in \cite{wueller2008statistic}:

$2^{EV}=\frac{f^2}{t}+\frac{ISO_{setting}}{ISO_{100}}$,

where $EV$ is exposure value, $f$ is the f-number, $t$ is the exposure time and $ISO_{setting}$ is the ISO used to take the photograph. From EV we can derive the illumination level (assuming that the photograph is properly exposed) as in \cite{wueller2018statistic}. We define low illumination between -4 and 7 EV (up to 320lx), mid-level illumination between 8 and 10 EV (640 to 2560lx) and high-level illumination above 11 EV (more than 5120lx) which approximately matches the setup in \cite{wu2017active}. Figure \ref{shutter_ISO_distributions} gives the distributions of exposure times (shutter speeds) in the COCO and  VOC2007 datasets. Table \ref{EV_num_samples} shows the image counts in each illumination level, not surpisingly, nearly 90\% of all images are acquired under high to medium illumination conditions.

\begin{table}[ht!]
\centering
\resizebox{0.9\columnwidth}{!}{%
\begin{tabular}{l|lll}
      & \multicolumn{3}{c}{Illumination Level}                                                      \\ \cline{2-4} 
      & \multicolumn{1}{l|}{High}          & \multicolumn{1}{l|}{Med}           & Low         \\ \hline
Train & \multicolumn{1}{l|}{42,045 (61\%)} & \multicolumn{1}{l|}{18,456 (27\%)} & 8054 (12\%) \\
Val   & \multicolumn{1}{l|}{1,832 (62\%)}  & \multicolumn{1}{l|}{771 (27\%)}    & 354 (12\%) 
\end{tabular}%
}
\caption{\label{EV_num_samples} Number of samples taken in high/medium/low illumination conditions in training and validation split in COCO dataset.}
\label{illum_num_samples}
\vspace{-1em}
\end{table}

 If our hypothesis that some imbalance in data distribution across sensing parameters might be the cause of the even performance was correct, then this should be revealed by examining algorithm performance across these sensor setting ranges.

\section{Object Detection on Images With Different Sensor Parameters from COCO Dataset}
We next investigated how different sensor parameters represented in those datasets affect the performance of object detection algorithms. We used several state-of-the art object detection algorithms, specifically, Faster R-CNN \cite{ren2015faster}, Mask R-CNN \cite{he2017mask}, YOLOv3 \cite{redmon2018yolov3} and RetinaNet \cite{lin2017focal}.  All are trained on COCO \texttt{trainval35K} set. Figures \ref{COCO_all_train} and \ref{COCO_all_val} show the percentages of images for a range of the shutter speed and ISO settings in COCO train and validation sets. The bin edges of heatmaps approximately match the ranges reported in \cite{andreopoulos2011sensor} and \cite{wu2017active}. Since shutter speeds of the cameras used in the previous works were limited to 1s, in our setup all images with exposure time $>1$s fall into the last bin and exposure time values between 0 and 1 are split into 8 equal intervals. Both \cite{andreopoulos2011sensor} and \cite{wu2017active} report gain, which is not available on most consumer cameras, therefore we use ISO values as a proxy. The following ISO bin ranges [0, 100, 200, 400, 800, 1600, 3200, 6400, 10000] approximately correspond to the gain values used in the previous works.

\begin{figure}[ht]
\centering
\begin{subfigure}[b]{0.48\columnwidth}
\includegraphics[width=\columnwidth]{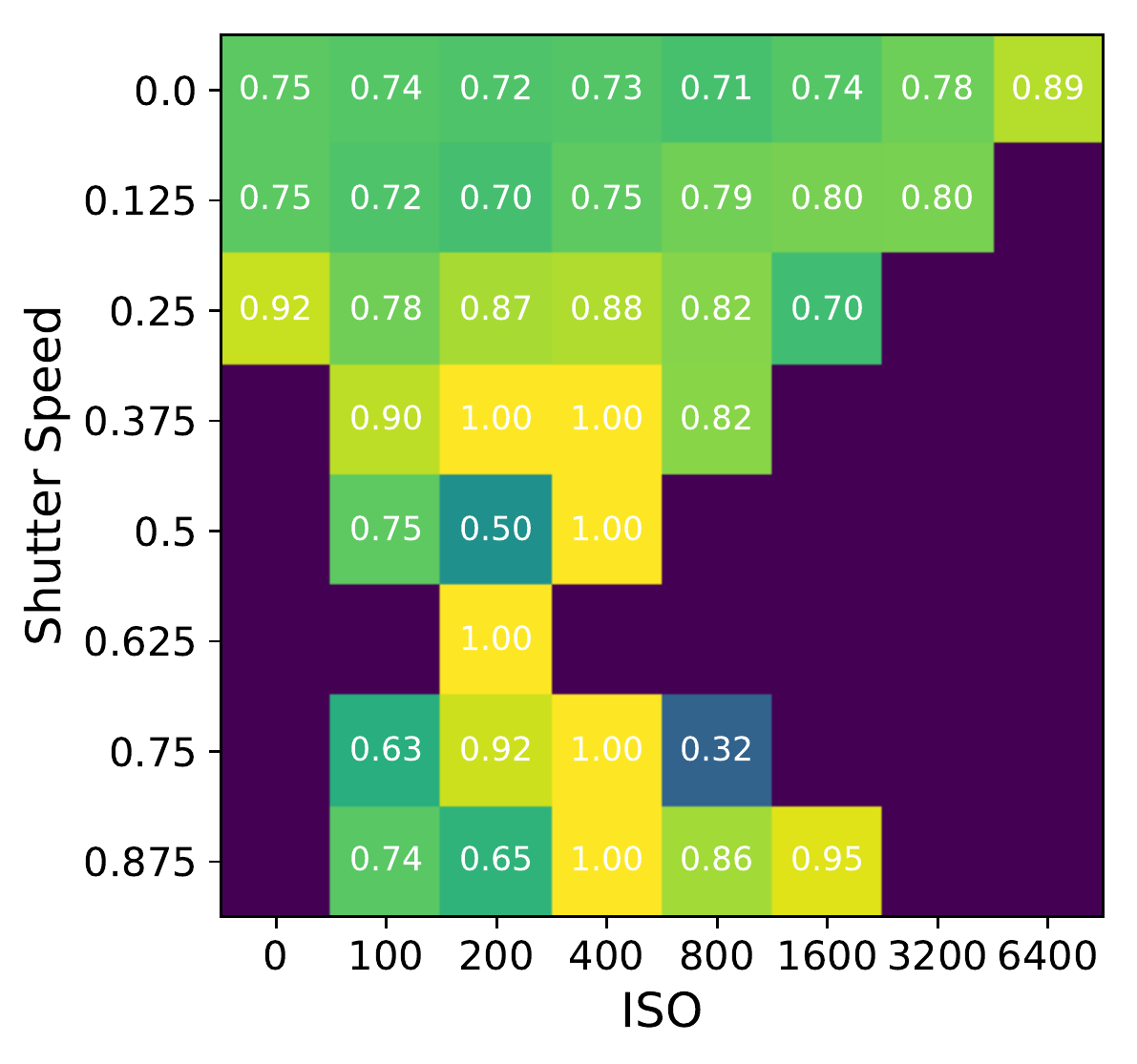}
\caption{YOLOv3}
\end{subfigure}
\begin{subfigure}[b]{0.48\columnwidth}
\includegraphics[width=\columnwidth]{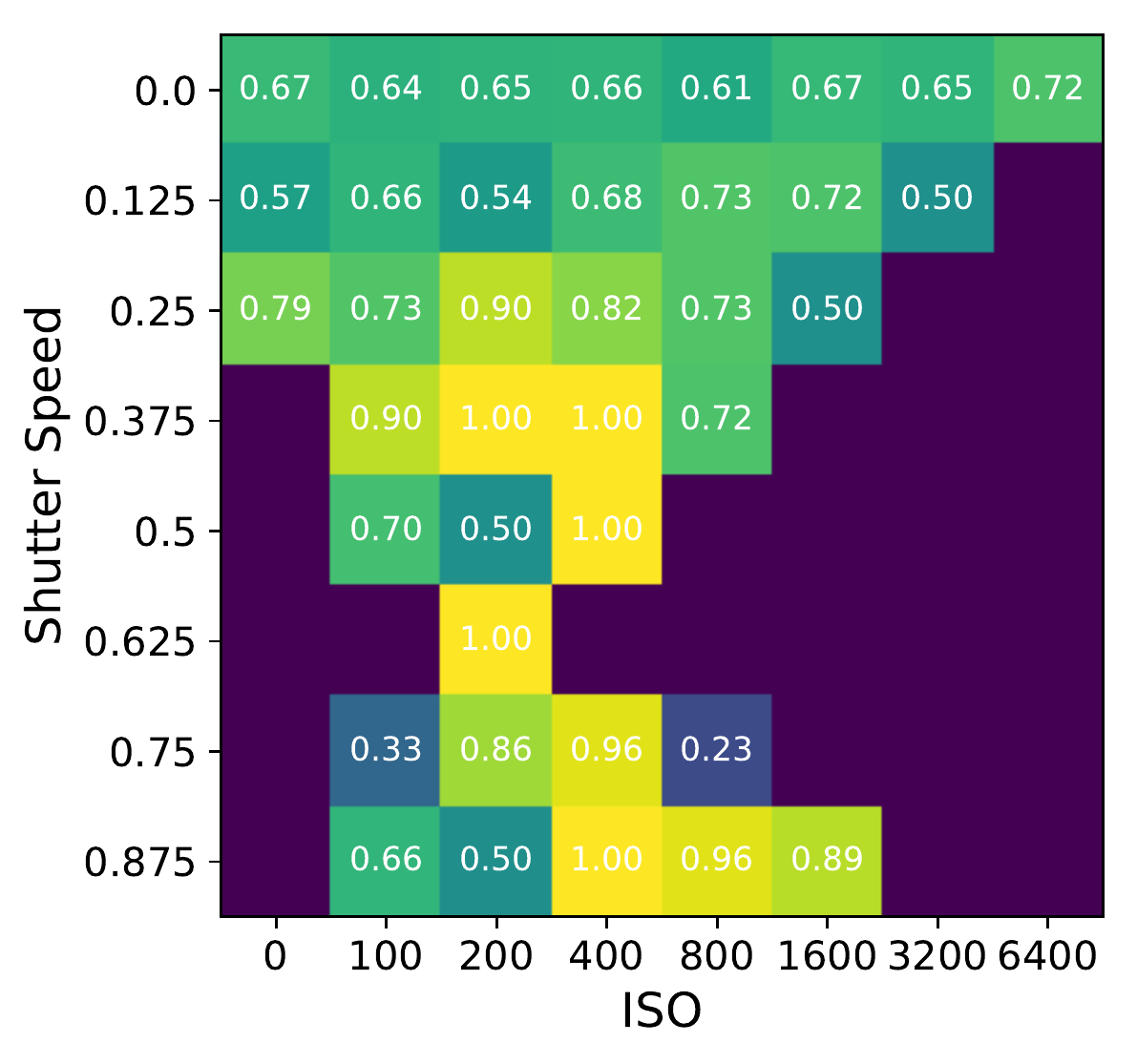}
\caption{Faster R-CNN}
\end{subfigure}
\begin{subfigure}[b]{0.48\columnwidth}
\includegraphics[width=\columnwidth]{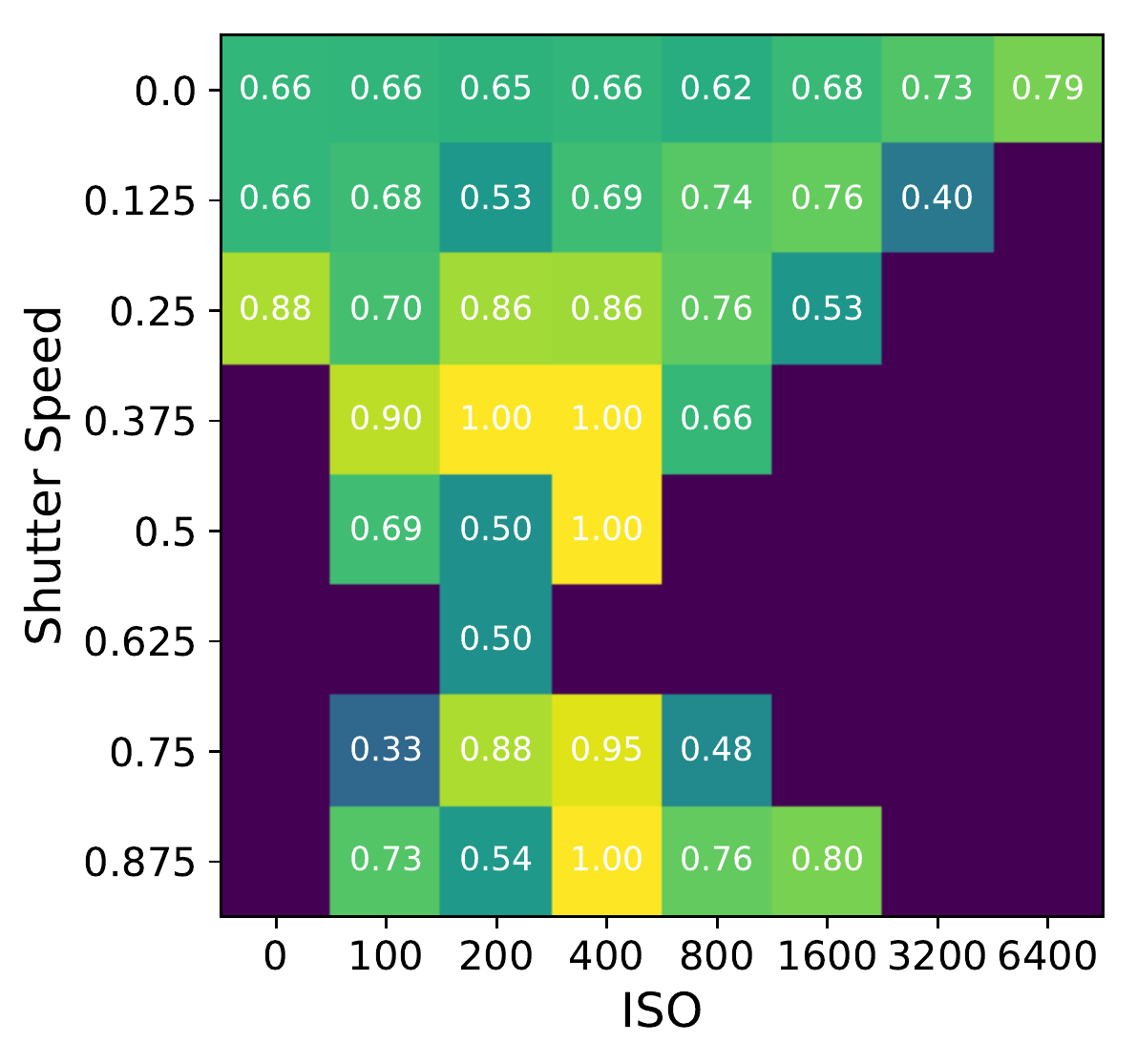}
\caption{Mask R-CNN}
\end{subfigure}
\begin{subfigure}[b]{0.48\columnwidth}
\includegraphics[width=\columnwidth]{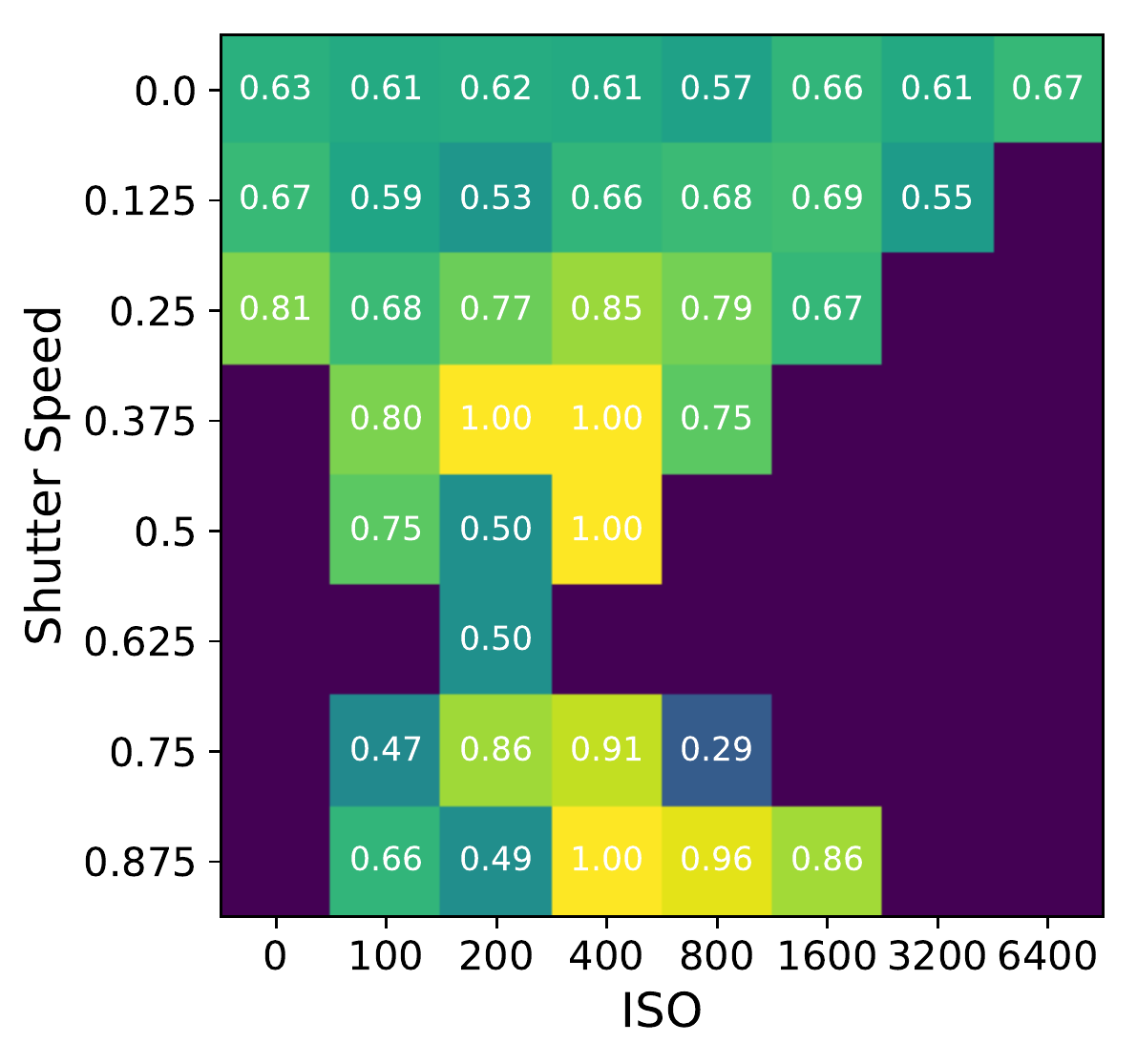}
\caption{RetinaNet}
\end{subfigure}
\caption{\label{COCO_eval}Results of evaluation of modern object detection algorithms on the \texttt{minival} subset of COCO for different sets of shutter speed and gain values. Mean average precision for bounding boxes at 0.5 IoU ($mAP^{IoU=.50}$) is computed for each bin using COCO API. }
\end{figure}

Figure \ref{COCO_eval} shows evaluation results in terms of mean average precision (mAP) for state-of-the-art object detection algorithms trained on COCO dataset and evaluated on the portion of COCO 5K \texttt{minival} set with available EXIF data and presented in the same style as the previous tests. It is difficult to compare our results with the results of the previous works directly because of the differences in the evaluation datasets, algorithms (interest point detection vs object detection), camera parameters (gain vs ISO), inability to precisely establish illumination level in common vision datasets and possible inconsistencies in computing average precision in each case. Nevertheless, a great deal can be observed. 

Note that nearly 90\% of training and validation data in COCO is concentrated in the top row of the diagram (very short exposure times and ISO values of up to 800). Figure \ref{COCO_eval} reveals very similar results from all 4 algorithms that are trained on this dataset suggesting possible training bias. It is also apparent that the mAP values in the top row are consistent with the reported performance of the algorithms but fluctuate wildly in bins that contain less representative camera parameter ranges. It is hard to attribute this fluctuation entirely to the sensor bias, as other factors may be at play (e.g. types and number of objects, small number of images in the underrepresented bins). This it is something that should be investigated further. It is never a useful property for an algorithm to display such significant sensitivity to small parameter changes. For imaging, one might expect a small shift in shutter speed, for example, to lead to only small changes in subsequent detection performance; this experiment shows this is not the case.

\section{Discussion}
The above experiments revealed several key points. First, theory-driven algorithms seem to have an orderly pattern of performance with respect to the sensor settings we examined. This may be due to their analytic definitions; they were not designed to be parameterized for the full range of sensor settings. One can easily conclude that if good performance is sought from any of these algorithms, they should be employed with cameras set to the algorithm's inherent optimal ranges. Of course, this limited test needs to be greatly expanded before finalizing this conclusion. However, in an engineering context this makes good sense: commercial products are commonly wrapped in detailed instructions about how to use and not use a product to ensure expected performance. In retrospect, knowledge of algorithm performance under varying sensor settings would have enhanced their design.

Second, the same test on modern algorithms reveals a haphazard performance pattern. It might be that some of the large variations are due to biases in the data, maybe some due to the particular objects in question, Others may be due to the properties of the network architectures (e.g., \cite{rosenfeld2018elephant}). This also needs much deeper analysis. Nevertheless, the performance pattern of Figure 5
is not entirely inconsistent with the training sample distributions of Figure \ref{dataset_sensor_bias}  and \ref{shutter_ISO_distributions}. The often seen claims of generalizable behavior for such algorithms would lead one to think that they might not be so sensitive to biases or other variations but they clearly are. Further, even though the performance shown in
Figure \ref{COCO_eval} effectively only uses about 60\% of the dataset as tests, the original system was trained on the full dataset.

Third, examination of two popular image sets, VOC2007 and COCO, shows that image metadata (sensor settings, camera pose, illumination, etc.) is often not available. This means that for any given `in the wild' set of images, the performance of data-driven methods may be predicted by how well the distribution of images along dimensions of sensor setting and illumination parameters in a test set matches the distribution resulting from the training set. This requires further verification. 

Fourth, and perhaps most revealing, we see that the comparison of high performing regions in Figure 1 with distribution of sensor settings of Figure 4 reveals only a tiny overlap: on those datasets, those algorithms would all perform terribly. This is not due to their bad design. It is due partly to their use outside of their design specifications, in this case, empirically obtained
camera sensor settings. More complete consideration of empirical methodology - where all aspects of the data are recorded and used to ensure reproducibility - might have led to different results in a head-to-head comparison.

This may also suggest that generalization of performance across image sets may not be a sensible expectation. In fact, the direct comparison of theory-driven vs data-driven methods was inappropriate because the theory-driven methods never had a chance due to the test set mis-matches just described. Any of the interest point algorithms in our first test could be applied to any image of either COCO or VOC2007 datasets and would perform terribly
only because the test ranges are not matched to
their inherent operating ranges. 

In some sense, the community has largely ignored sensor settings and their relevance as described here. This may be due to the belief that since humans can be shown images from any source, representing any sort of object or scene, familiar or unfamiliar, and provide reasonable responses as to their content, computer vision systems should also. This goal is ``baked in'' to the overall research enterprise and variations are considered either as nuisance or good tests of generalization ability. As an example of the latter, there are a number of studies investigating the effects of various image degradations on the performance of the deep-learning based algorithms. These include both artificially distorted images \cite{rodner2016fine, geirhos2018generalisation, roy2018effects, che2019gazegan} as well as more realistic transformations, for example, viewing the image through a cheap webcamera \cite{tow2016robustness}.

However, the proposition itself is not well-formed. It is simply untrue that humans, when shown an image out of the vast range of possibilities, perform in the same manner with respect to time to respond, accuracy of response, manner of response, and internal processing strategy. The behavioral literature solidly proves this. Even the computational literature points to this. While humans have been shown to be more robust than state-of-the-art deep networks to nearly all image degradations examined, human performance still gradually drops off as the levels of distortions increase \cite{dodge2017study, geirhos2018generalisation}.

As we mentioned earlier, most of the experiments investigating the robustness of the deep networks to image distortions do not tie those to the changes resulting from camera settings. Perhaps changing ISO from 100 to 200, and the resulting increase in noise, would not be easily noticeable to human eye, but may have a measurable effect on the performance of CNNs which are known to be affected by very small changes in the images (e.g. small affine transformations\cite{azulay2018deep}, various types of noise \cite{geirhos2018generalisation}). As \cite{su2019one} shows changing a single pixel in the right place is all it takes sometimes. It should be clear that the ``baked in'' belief  needs much more nuance and a significantly deeper analysis, if not outright rejection.

Finally, as can be seen in Figures \ref{dataset_sensor_bias} and \ref{COCO_eval}, the variability required to train is not even available in the large datasets we considered. The distributions of images across these parameters was very uneven so training algorithms are impeded with respect to learning the variations. It might be a good practice to require similar demonstrations of image distributions across the relevant parameters in order to ensure that no only training, but comparative evaluations, are propery performed.

\section{Conclusions}

With all due respect to the terrific advances made in computer vision by the application of deep learning methods - here termed data-driven models - we propose here, and provide some justification, that the empirical methodology that led to the turning point in the discipline was based on an oversight that none of us noticed at the time. Sensor settings matter and each algorithm, perhaps most especially the theory-driven ones, have ranges within which one might expect good performance and ranges where one should not expect it. Testing outside the ranges is not only unfair but completely inappropriate. Theory-driven algorithms were compared against data-driven algorithms on datasets unrepresentative of the theory-driven algorithm operating ranges but on which the data-driven algorithms were trained. 

If sensor settings (maybe also illumination levels or other imaging aspects) played a role in the large scale testing of the classic theory-driven algorithms, then perhaps they would have performed at high levels. The community did not do this. In comparing the data-driven with theory-driven algorithms the distribution of camera settings (unknown to everyone) favored the data-driven algorithms because they were trained on such a random distribution while the theory-driven algorithms were tested on data for which they were not designed to operate. But no one realized this at the time. Thus the empirical strategy favored data-driven models.

If the empirical strategy specifically included a fuller specification of optimal operating ranges for all algorithms and each algorithm were tested accordingly, what would their performance rankings be? The theory-driven algorithms would have been tested only on images from camera settings for which they were
designed and perhaps the performance gap would have been smaller. Such greater specificity in experimental design seems more common in other disciplines. An empirical method involves the use of objective, quantitative observation in a systematically controlled, replicable situation, in order to test or refine a theory. The key features of the experiment are control over variables (independent, dependent and extraneous), careful objective measurement and establishing cause and effect relationships. At the very least, a
discussion on how to firm up empiricism in computer vision needs to take place.

The impact of this is great. Knowledge of the sensor setting ranges that lead to poor performance would allow one to ensure failure of
a particular algorithm. This is as true for any head-to-head technical competition as it is for an adversary. Non-linear performance of any algorithm along one or more parameter settings is a highly undesirable characteristic. It may be that a blend of the approaches, data-driven and theory-driven, would permit better specification towards providing linear performance across the relevant parameters but within a framework where learning can occur using datasets that better represent the target population.

\vspace{-0.4em}
\subsection*{Acknowledgements} This research was funded by the Natural Sciences and Engineering Research Council of Canada, the Canada Research Chairs Program and the Air Force Office for Scientific Research. Part of this work (sections 2 and 3) was performed when AA and YW were graduate students in the Tsotsos Lab. The authors thank Konstantine Tsotsos who provided useful comments and suggestions.
{\small
\bibliographystyle{ieee}
\bibliography{arxiv}
}

\end{document}